\documentclass[sigconf,natbib=false]{acmart}
\AtBeginDocument{%
  }

\usepackage{listings}
\usepackage{graphicx} 

\usepackage{framed}
\usepackage{xcolor}
\usepackage{multirow}
\usepackage{multicol}
\usepackage{enumitem}
\usepackage{array}
\usepackage{booktabs}  
\usepackage{tabularx}

\newcounter{customboxcounter}
\newenvironment{custombox}[1]{
  \refstepcounter{customboxcounter}
  \MakeFramed{\FrameRestore}
  \noindent\textbf{#1} 
  \par\vskip6pt}{
  \endMakeFramed}

\setcopyright{none}
\copyrightyear{2025}
\acmYear{2025}
\acmDOI{}
\acmConference[CLAIRvoyant 2024 and CLAIRvoyantS 2025 at ICAIL 2025]{2nd ConventicLe on Artificial Intelligence Regulation and Safety at ICAIL 2025}{June 16,
  2025}{Chicago, IL}
\acmISBN{}

\RequirePackage[
  datamodel=acmdatamodel,
  style=acmnumeric,
  ]{biblatex}

\begin{document}

\title[A Reflective Multi-Agent Approach for Legal Argument Generation]{Mitigating Manipulation and Enhancing Persuasion: A Reflective Multi-Agent Approach for Legal Argument Generation}

\author{Li Zhang}
\orcid{0000-0003-0375-1793}
\authornotemark[1]
\affiliation{%
  \institution{Intelligent Systems Program \\ University of Pittsburgh}
  \city{Pittsburgh}
  \state{Pennsylvania}
  \country{USA}
}
\email{liz239@pitt.edu}

\author{Kevin D. Ashley}
\orcid{0000-0002-5535-0759}
\affiliation{%
  \institution{Intelligent Systems Program \\ University of Pittsburgh}
  \city{Pittsburgh}
  \state{Pennsylvania}
  \country{USA}
}
\email{ashley@pitt.edu}

\renewcommand{\shortauthors}{Li et al.}

\begin{abstract}
Large Language Models (LLMs) are increasingly explored for legal argument generation, yet they pose significant risks of manipulation through hallucination and ungrounded persuasion, and often fail to utilize provided factual bases effectively or abstain when arguments are untenable. This paper introduces a novel reflective multi-agent method designed to address these challenges in the context of legally compliant persuasion. Our approach employs specialized agents (factor analyst and argument polisher) in an iterative refinement process to generate 3-ply legal arguments (plaintiff, defendant, rebuttal). We evaluate reflective multi-agent against single-agent, enhanced-prompt single-agent, and non-reflective multi-agent baselines using four diverse LLMs (GPT-4o, GPT-4o-mini, Llama-4-Maverick-17b-128e, Llama-4-Scout-17b-16e) across three legal scenarios: ``arguable'', ``mismatched'', and ``non-arguable''. Results demonstrate that the reflective multi-agent approach excels at successful abstention by preventing generation when arguments cannot be grounded, improves hallucination accuracy by reducing fabricated and misattributed factors and enhances factor utilization recall by better using the provided case facts. These findings suggest that structured reflection within a multi-agent framework offers a robust method for fostering ethical persuasion and mitigating manipulation in LLM-based legal argumentation systems.

\hspace{1em}

\textbf{Project Page:} \href{https://lizhang-aiandlaw.github.io/A-Reflective-Multi-Agent-Approach-for-Legal-Argument-Generation/}{lizhang-aiandlaw.github.io/A-Reflective-Multi-Agent-Approach-for-Legal-Argument-Generation}

\end{abstract}





\maketitle

\section{Introduction}

Argumentation, at its core, is the art of persuasion, employing logical reasoning and evidence to influence an audience on a particular matter \cite{nickerson2020argumentation}. In the legal domain, argumentation takes on a specialized form, where the careful construction of evidence-based claims is crucial. The ability to persuade effectively, while adhering to ethical and factual standards, is fundamental to legal practice \cite{rhode1985ethical}. As LLMs become increasingly refined, their potential to assist in, and even automate, aspects of legal argument generation brings both immense opportunities and challenges, particularly concerning the integrity of persuasion in this high-stakes field.

\subsection{The Rise of LLMs in Legal Domain}
The integration of LLMs into the legal domain presents a paradigm shift, offering considerable potential for automating routine tasks, assisting in legal research, document drafting, and even argument generation \cite{siino2025exploring,gray2025generating,zhang2025measuring}. LLMs can swiftly sift through vast legal data, draft legal documents, and assist in complex legal analysis, thereby enabling lawyers to focus on more strategic aspects of their work \cite{nielsen2024building}. However, the inherent persuasive capabilities that make these models valuable also introduce a complex dichotomy: while they can be leveraged for beneficial applications, they simultaneously pose risks of manipulation and unethical influence if not carefully managed \cite{bommasani2021opportunities}. The primary challenge lies in harnessing their power for ethical assistance while mitigating the potential for misuse.

These risks are amplified by the fact that the persuasive influence of LLMs is not unidirectional. Beyond their role as persuaders, these systems can also be susceptible to persuasion themselves, rendering them vulnerable to adversarial attacks and the reinforcement of biases present in their training data or input prompts \cite{bozdag2025readsystematicsurveycomputational}. For instance, an LLM tasked with generating legal arguments might have been trained on datasets containing systemic biases or could be influenced by subtly crafted adversarial inputs during its operational lifecycle \cite{draper2023potential}. This ``persuadee'' vulnerability means that an LLM could inadvertently generate arguments that, while appearing coherent and persuasive, perpetuate these biases or reflect manipulated information, thereby subtly distorting legal outcomes. Such complex interplay between an LLM's persuasive output and its susceptibility to influence necessitates robust internal validation mechanisms within LLM-based legal chatbots. These mechanisms must go beyond mere output generation to meticulously scrutinize the factual grounding, logical consistency, and potential biases of the arguments produced, ensuring their integrity and reliability.

\subsection{The Triad of Challenges}
Pilot studies and broader research have identified shortcomings in LLM performance when applied to legal domain including argument generation \cite{dahl2024large,gray2025generating,zhang2025measuring}. Three challenges emerge: hallucination, inadequate abstention, and poor factor utilization.

Hallucination, the generation of text that is factually incorrect, inconsistent, or not supported by input data, is particularly pernicious in the legal field where precision and accuracy are fundamental. Studies have reported alarmingly high hallucination rates by LLMs in response to legal queries, with models fabricating case details or misstating legal principles \cite{magesh2024hallucination}. Such inaccuracies can lead to nonsensical or harmful legal advice \cite{krook2024large}.

Abstention, the ability of a model to refrain from answering when an argument is ungroundable or when it lacks sufficient information, is crucial for reliability. However, LLMs often fail to recognize the boundaries of their knowledge or the untenability of a query, proceeding to generate responses even when they should abstain \cite{wen2024know}.

Factor utilization refers to the extent to which an LLM incorporates relevant factual elements (factors) from provided case materials into its generated arguments \cite{nadler1983evidence}. Poor factor utilization results in arguments that may be superficially plausible but lack substantive grounding in the specific facts of the case, thereby diminishing their persuasive strength and utility.

These three challenges are often interlinked. A failure to abstain in scenarios where an argument cannot be legitimately grounded (``mismatched'' or ``non-arguable'' cases) increases the likelihood of hallucination. If a model is compelled to generate an argument without a proper factual basis, it is more prone to inventing facts or misapplying existing ones to fulfill the generation task. This directly translates to a higher risk of manipulation, as the generated argument may appear coherent but be based on falsehoods or irrelevant information.

\subsection{Our Contribution: A Reflective Multi-Agent Approach}
To address this triad of challenges, this paper introduces a Reflective Multi-Agent method for generating 3-ply legal arguments. The Reflective Multi-Agent framework is designed to enhance ethical persuasion by improving factor utilization and grounding, while reducing manipulation by minimizing hallucinations and promoting appropriate abstention. The core of the Reflective Multi-Agent approach lies in its utilization of specialized LLM-based agents—a Factor Analyst and an Argument Polisher—which engage in an iterative reflection and refinement process for each ply of the argument.

\subsection{Guiding Research Questions}
This research is guided by the following questions:
\begin{itemize}
    \item \textbf{RQ1:} How does the proposed reflective multi-agent approach compare to single-agent, enhanced-prompt single-agent, and non-reflective multi-agent methods in terms of (a) hallucination accuracy, (b) factor utilization recall, and (c) successful abstention ratio when generating 3-ply legal arguments?
    \item \textbf{RQ2:} What is the impact of the reflection mechanism on the quality of generated arguments across different LLMs and varying legal scenario complexities (arguable, mismatched, non-arguable)?
    \item \textbf{RQ3:} To what extent can the Reflective Multi-Agent framework contribute to the development of more ethically persuasive and less manipulative legally compliant intelligent chatbots?
\end{itemize}

\subsection{Overview of Contributions and Manuscript Roadmap}
The primary contributions of this work include: (i) the design and implementation of a novel Reflective Multi-Agent framework for legal argument generation; (ii) an empirical evaluation of the Reflective Multi-Agent framework against several baseline methods using multiple LLMs across diverse and challenging legal scenarios; and (iii) an analysis that connects the technical performance improvements to the broader goals of achieving ethical persuasion and mitigating manipulation in legally compliant intelligent chatbots.

The remainder of this paper is structured as follows: Section 2 reviews related work. Section 3 details the architecture and workflow of the proposed Reflective Multi-Agent framework. Section 4 describes the experimental design. Section 5 presents and analyzes the experimental results. Section 6 discusses the implications. Section 7 outlines limitations and future research. Finally, Section 8 concludes the paper.

\section{Related Work}
The pursuit of reliable and ethically sound AI systems for legal argument generation intersects with several active research domains. This section reviews pertinent literature on the application of LLMs in legal contexts, computational models of persuasion and manipulation, the role of multi-agent systems in complex reasoning, the development of reflective and self-correcting mechanisms in LLMs, and methodologies for evaluating their outputs, particularly concerning hallucination, abstention, and content grounding.

\subsection{LLMs for Legal Reasoning and Argument Generation}
The application of LLMs to legal tasks, including argument generation, is growing \cite{siino2025exploring,gray2025generating,zhang2025measuring}. Computational legal argument models, especially those using case-based reasoning and `factors' (stereotypical fact patterns), offer a strong foundation for this endeavor, with systems like HYPO pioneering factor-based case analysis \cite{ashley1990}, followed by developments in factor hierarchies (CATO \cite{aleven1997teaching}) and value incorporation \cite{grabmair2017predicting}. These factor-based methods are key for evaluating argument factuality, central to our study. However, LLMs face challenges in specialized legal argument construction \cite{padiu2024extent} and are prone to ``legal hallucinations''—generating factually incorrect or misapplied legal content \cite{dahl2024hallucinating}—highlighting the need for domain-specific adaptations and robust verification, as addressed by our Reflective Multi-Agent framework, which aims to enhance the grounding and accuracy of LLM-generated legal arguments, building on prior work to enhance LLM performance in law.

\subsection{Computational Models of Persuasion and Manipulation in AI}
The persuasive capabilities of AI systems, particularly LLMs, are increasingly recognized, with models functioning as persuaders, being susceptible to persuasion themselves, and even acting as arbiters of persuasive attempts \cite{breum2024persuasive}. This multifaceted role presents opportunities for beneficial applications but concurrently introduces risks of manipulation, social engineering, and unethical influence \cite{hu2025j}. A central challenge, especially within sensitive domains such as law, is the development of ``computable methods for ensuring ethical persuasion'' \cite{rogiers2024persuasion}.
The concept of ``computational manipulation'' is particularly germane to legal AI. As explored in contexts like AI-driven persuasive technology in contract law, AI systems can exploit user data and cognitive biases to influence decisions, potentially undermining autonomous decision-making processes without explicit user awareness \cite{faraoni2024contract,burtell2023artificial}. This notion directly correlates with problematic LLM behaviors, such as the generation of ``favorable'' yet false factual assertions or the vigorous defense of an untenable position. Our work seeks to mitigate such manipulative tendencies by enhancing factual grounding and promoting abstention when arguments lack merit.

\subsection{Multi-Agent Systems in Complex Reasoning and Dialogue}
LLM-based Multi-Agent Systems (MAS) are emerging as a potent paradigm for addressing complex problems, leveraging distributed task handling and specialized agent roles to enhance robustness and reasoning capabilities \cite{guo2024large,gao2024agentscope}. Multi-agent debate frameworks, for instance, have been proposed to improve the quality of reasoning by fostering interaction among agents, thereby potentially counteracting issues like the ``Degeneration-of-Thought'' problem observed in single-agent systems \cite{liang2023encouraging}.
The proposed Reflective Multi-Agent system builds upon these principles by employing distinct agents for Plaintiff and Defendant roles. Furthermore, it introduces specialized sub-agents—the Factor Analyst and Argument Polisher—which collaborate to meticulously refine arguments. This architecture represents a nuanced application of Multi-Agent Systems, designed to harness the benefits of distributed expertise and structured dialogue for improved legal argument generation.

\subsection{Reflection, Self-Correction, and Iterative Refinement in LLMs}
Mechanisms enabling reflection and self-correction are crucial for enhancing the reliability and accuracy of LLM outputs \cite{pan2023automatically}. However, the efficacy of unaided self-correction remains a subject of debate. The ``SELF-[IN]CORRECT'' hypothesis, for example, posits that LLMs may not be consistently better at discriminating the quality of their own responses than they are at generating initial ones \cite{huang2023large}. This suggests that structured and guided reflection mechanisms may prove more effective. Challenges inherent in self-correction, such as difficulties in error detection and inherent self-bias, further motivate the development of more robust and systematic approaches \cite{xi2024enhancing}.
The Reflective Multi-Agent framework's reflective process is designed to be explicit and role-based, employing specialized agents (the Factor Analyst and Argument Polisher) for distinct analytical and refinement tasks. This structured critique aims to overcome the limitations associated with unguided self-correction, which demonstrates the power of structured reflection in improving LLM performance \cite{shinn2023reflexion}.

\subsection{Metrics for Evaluating LLM Outputs}
The rigorous evaluation of LLM-generated content, particularly in high-stakes domains like law, necessitates robust and domain-relevant metrics. For assessing hallucination, various benchmarks and detection methods are continuously being developed \cite{ji2023survey,luo2024hallucination,ravi2024lynx}. The distinction between extrinsic and intrinsic hallucination, as offered by frameworks like ``HalluLens'' \cite{bang2025hallulens}, is particularly relevant; our work primarily addresses intrinsic hallucination by ensuring arguments are grounded in provided case factors.
The capacity for appropriate abstention is another aspect of LLM reliability \cite{feng2024don}. Surveys and studies provide comprehensive overviews of abstention methods and evaluation metrics, highlighting the importance of this capability \cite{wen2024know,madhusudhan2024llms}.
Furthermore, evaluating factor utilization, or the faithfulness of generated arguments to source material, is essential for ensuring the substantive quality of legal outputs \cite{liu2023trustworthy}. Frameworks like ICAT, which evaluate factual accuracy and coverage by decomposing text into claims and aligning them with relevant aspects (factors), offer valuable methodologies \cite{samarinas2025beyond}. QA-based verification methods \cite{fabbri2021qafacteval} share conceptual similarities with our approach of employing an external LLM for factor summarization and comparison. This aligns with the broader trend of using ``LLM-as-a-Judge'' approaches for nuanced evaluations, a field with its own set of evolving best practices and considerations \cite{zheng2023judging,gu2024survey}. Our evaluation methodology draws upon these established principles to provide an assessment of the Reflective Multi-Agent framework.

\section{The Reflective Multi-Agent Framework}
\subsection{Architectural Overview}
\begin{figure}[h]
    \includegraphics[width=\linewidth]{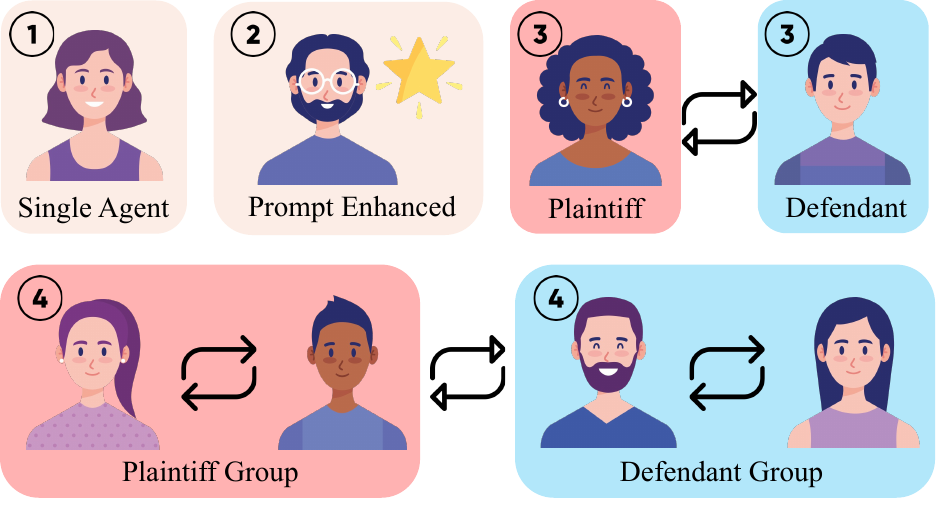}
    \caption{Overview of the Agentic Structure for Legal Argument Generation, including the RMA framework's reflective components (Factor Analyst, Argument Polisher) interacting with the Argument Developer.}
    \label{fig:overview_structure}
\end{figure}

\begin{figure}[h]
    \centering
    \includegraphics[width=\linewidth]{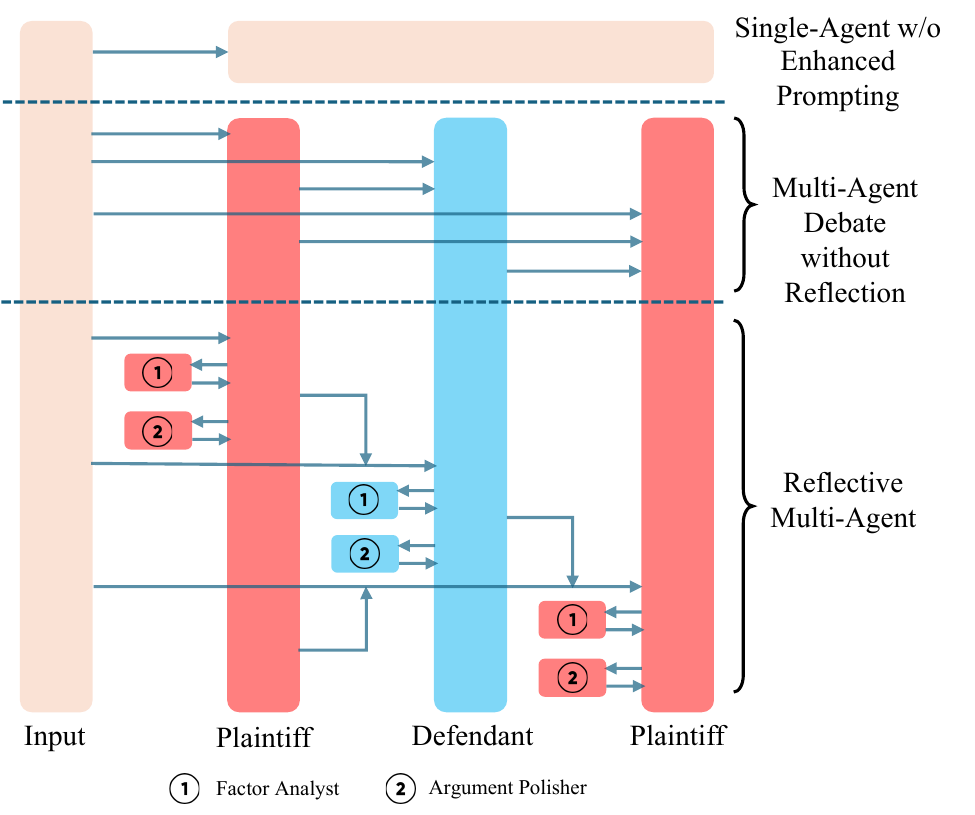}
    \caption{Information Flow of Different Structures: Single Agent (SA), Single Agent with Enhanced Prompting (SA-EP), Multi-Agent Debate without Reflection (MA), and Reflective Multi-Agent (RMA).}
    \label{fig:information_flow}
\end{figure}

The Reflective Multi-Agent framework is designed to generate a 3-ply legal argument structure \cite{ashley1990,zhang2025measuring}:
\begin{enumerate}
    \item The Plaintiff's initial argument.
    \item The Defendant's counterargument.
    \item The Plaintiff's rebuttal to the Defendant's counterargument.
\end{enumerate}
The core argument generation and reflection process is applied sequentially to produce each ply, with context from previous plies maintained for coherence. The overall agentic structure and information flow for different configurations, including Reflective Multi-Agent, are depicted in Figure \ref{fig:overview_structure} and Figure \ref{fig:information_flow}.


\subsection{Agent Roles}
The reflection mechanism is driven by two specialized LLM-based agents:

\textbf{Factor Analyst:}
\begin{itemize}
    \item \textit{Purpose:} Substantive review of the generated argument.
    \item \textit{Functions:} Detect Hallucinations (identify factors not in input cases c1, c2, c3), Mandate Abstention (output ``TERMINATE'' if argument is untenable).
    \item \textit{Input:} Generated argument for the current ply, original input factors (c1, c2, c3).
    \item \textit{Output:} Analysis report (hallucinations, revision/accept/terminate decision).
    \item \textit{Prompting Strategy:} ``You are a Factor Analyst. Review the following Argument based only on the provided Case Factors (c1, c2, c3). Identify any statements or factors in the Argument that are not supported by c1, c2, or c3. List all factors from c1, c2, and c3 that were utilized. If the combination of c1, c2, and c3 does not provide a basis for a valid argument for the current side due to lack of overlapping relevant factors or mismatched outcomes, your primary output must be 'TERMINATE'. Otherwise, provide your analysis.'' 
\end{itemize}

\textbf{Argument Polisher:}
\begin{itemize}
    \item \textit{Purpose:} Rhetorical and stylistic quality of the argument.
    \item \textit{Functions:} Assess Factor Utilization, Enhance Clarity and Coherence, Improve Persuasiveness, Adhere to Legal Style.
    \item \textit{Input:} Original/revised argument, Factor Analyst's report.
    \item \textit{Output:} Polished suggestions.
    \item \textit{Prompting Strategy:} ``You are an Argument Polisher. Refine the following Argument to enhance its clarity, coherence, and persuasive impact. Ensure all claims remain strictly grounded in the factors identified and approved by the Factor Analyst. Address the following specific points if provided: [points from Factor Analyst's report]. Do not introduce new factual claims or alter the core factual basis.''
\end{itemize}

\subsection{The Iterative Reflection Workflow}
The Reflective Multi-Agent framework employs a structured iterative workflow for each ply:
\begin{enumerate}
    \item \textbf{Initial Argument Generation:} Primary LLM generates an initial argument.
    \item \textbf{Round 1 Analysis \& Polishing:} Factor Analyst assesses, then Argument Polisher refines.
    \item \textbf{Revision Decision \& Execution:} If Polisher deems revision necessary (or Factor Analyst mandates critical changes), the primary LLM revises argument once based on consolidated feedback.
    \item \textbf{Round 2 Analysis \& Polishing (if revised):} Factor Analyst re-checks, Argument Polisher re-polishes.
    \item \textbf{Final Output for Ply:} ``TERMINATE'' if mandated by Factor Analyst. Otherwise, the polished initial argument (if no revision) or re-polished revised argument.
\end{enumerate}
This sequence repeats for Defendant's Counterargument and Plaintiff's Rebuttal. This structured deliberation with bounded iteration balances reflection benefits with efficiency.

\subsection{Comparative Baselines}
\begin{itemize}
    \item \textbf{3.4.1. Single Agent (SA):} Basic configuration. Single LLM generates 3-ply argument in one turn with standard prompts.
    \item \textbf{3.4.2. Single Agent with Enhanced Prompting (SA-EP):} Single LLM with Chain-of-Thought (CoT) style instructions focusing on minimizing hallucination, abstaining when ungroundable, and maximizing factor utilization.
    \item \textbf{3.4.3. Multi-Agent Debate without Reflection (MA):} Two LLM agents (Plaintiff, Defendant) generate arguments sequentially, with access to previous arguments, but no explicit Factor Analyst or Argument Polisher, nor iterative reflection. Analogous to basic debate structures.
\end{itemize}

\section{Experimental Design}
\subsection{Task Definition}
The core task is to generate a 3-ply legal argument. The input are three cases (c1: current case, c2: Plaintiff's precedent, c3: Defendant's precedent) represented by legal factors. These legal factors are derived from foundational work in legal AI focusing on U.S. trade secret misappropriation law \cite{ashley1990}. A standardized set of 26 factors is used, encapsulating key factual aspects of a case, such as circumstances surrounding disclosure, security measures implemented, characteristics of the information, and relevant employee conduct. Each factor is designated as typically favoring either the plaintiff (P) or the defendant (D). For instance, a case might be represented as:

\begin{quote} 
\texttt{[Case Name] [Outcome] [Factors: F1 Disclosure-in-negotiations (D), F4 Agreed-not-to-disclose (P), F6 Security-measures (P)]} 
\end{quote}

This structured factor representation provides the essential ground truth for subsequent automated evaluation by allowing objective comparison between cases based on shared and distinguishing factors. The process follows a structure established by \cite{ashley1990}.
First, the system, acting as the Plaintiff, argues for a favorable outcome by citing one of the provided precedent cases (c2 or c3) as analogous, emphasizing shared pro-plaintiff factors between the current case (c1) and the cited precedent.
Second, the system takes on the role of the Defendant. It responds by distinguishing the precedent case cited by the Plaintiff (highlighting differing factors) and then presents the other precedent case as a counterexample that favors the defendant, focusing on shared pro-defendant factors between c1 and this counterexample precedent.
Third, the system acts again as the Plaintiff to deliver a rebuttal. This involves distinguishing the counterexample case cited by the Defendant and reinforcing the original argument by emphasizing factors that differentiate the current case from the Defendant's chosen precedent.
The expected output is either the structured Plaintiff's Argument (based on c1 and c2), the Defendant's Counterargument (based on c1 and c3), and the Plaintiff's Rebuttal (based on c1 and c2), or a ``TERMINATE'' signal if the system determines that a valid argument cannot be constructed based on the provided factors.

For the argument generation, a \texttt{``max\_tokens''} limit of 1,000 was set, which proved sufficient for the typical length of a 3-ply argument structure. Other standard parameters included \texttt{``top\_p''}=1, \texttt{``frequency\_penalty''}=0, and \texttt{``presence\_penalty''}=0. The factor extraction step, performed by the external evaluator LLM (\texttt{GPT-4.1}), also utilized fixed deterministic settings (\texttt{``temperature''}=0, \texttt{``top\_p''}=1) to ensure consistency in the evaluation process itself.

\subsection{Scenarios: Arguable, Mismatched, and Non-Arguable Case Triples}
Three scenarios based on factor relationships in c1, c2, c3:
\begin{itemize}
    \item \textbf{Arguable:} Genuine grounds for argumentation. c1 shares relevant overlapping factors with c2 (for Plaintiff) and c3 (for Defendant). Models should generate substantive arguments.
    \item \textbf{Mismatched:} Outcomes of c2/c3 conflict with their intended use (e.g., c2 for Plaintiff has unfavorable outcome). Models should abstain.
    \item \textbf{Non-Arguable:} No relevant factor overlaps between c1 \& c2, and c1 \& c3. Models should abstain.
\end{itemize}
``Mismatched'' and ``non-arguable'' scenarios test ethical AI behavior; abstention is key. An example of case structures for these scenarios is in Table \ref{tab:case_set_examples}.

The generation process for these case triples is parametric, allowing for the specification of several key aspects. These include the total number of case triples to generate, the complexity level (which dictates the number of factors assigned to each case, typically ranging from complexity-1 to complexity+1), and, crucially, the scenario ``mode'' (Arguable, Mismatched, or Non-Arguable). For this study, we generated sets of 90 case triples for each of the three scenarios, with a specified complexity level of 5. This means each case within a triple typically contained between 4 and 6 factors.

\begin{table*}[htbp]
\caption{Examples of Different Scenario Modes}
\label{tab:case_set_examples}
\centering
\small
\begin{tabular}{p{0.15\textwidth}p{0.25\textwidth}p{0.25\textwidth}p{0.25\textwidth}}
\toprule
\textbf{Mode} & \textbf{Current Case (c1)} & \textbf{Plaintiff Precedent (c2)} & \textbf{Defendant Precedent (c3)} \\
\midrule
Arguable & F4(P)*, F5(D)†, F23(D)  & outcome: Plaintiff\newline F2(P), F4(P)*, F16(D) & outcome: Defendant\newline F2(P), F5(D)†, F12(P)\\
\midrule
Mismatched & F4(P)*, F5(D)†, F23(D) & outcome: Defendant\newline F2(P), F4(P)*, F16(D) & outcome: Plaintiff\newline F2(P), F5(D)†, F12(P) \\
\midrule
Non-arguable & F6(P), F22(P) & outcome: Plaintiff\newline F1(D), F27(D) & outcome: Defendant\newline F16(D), F24(D) \\
\bottomrule
\end{tabular}
\vspace{-2mm}
\begin{minipage}{\textwidth}
\scriptsize Notes: Common factors between Current Case and Plaintiff Precedent (c2) marked with *; common factors between Current Case and Defendant Precedent (c3) marked with †. Factor representations are illustrative.
\end{minipage}
\end{table*}

\subsection{Models Evaluated}
Four LLMs were used:
\begin{itemize}
    \item GPT-4o (OpenAI)
    \item GPT-4o-mini (OpenAI)
    \item Llama-4-Maverick-17b-128e-instruct (Meta)
    \item Llama-4-Scout-17b-16e-instruct (Meta)
\end{itemize}
These cover a spectrum of capabilities (proprietary and open-source, comparatively small and large).

\subsection{Evaluation Metrics}
Metrics assess hallucination, factor utilization, and abstention. An external LLM (\texttt{GPT-4.1}) is used for automated factor extraction from generated arguments, comparing them to ground-truth input factors. Let $F_{Ext, c}$ be factors extracted by the evaluator for case $c \in \{c1, c2, c3\}$, and $F_{GT, c}$ be ground-truth factors for case $c$.

\textbf{4.4.1. Hallucination Accuracy:}
Quantifies avoidance of factual inaccuracies. A hallucinated factor is one mentioned in the argument for a specific case but not present in its ground-truth input.
The hallucination accuracy ($Acc_H$) is given by:
\begin{equation}
Acc_H = \left(1 - \frac{N_{h}}{N_{gt}}\right) \times 100\%
\end{equation}
where $N_{h}$ is the count of hallucinated factors, defined as:
\begin{multline} 
N_{h} = \sum_{c \in \{c1, c2, c3\}} \left| \{ f \in F_{Ext, c} \mid f \notin F_{GT, c} \} \right|
\end{multline}
and $N_{gt}$ is the total count of ground-truth factors from the input cases:
\begin{equation}
N_{gt} = \sum_{c \in \{c1, c2, c3\}} |F_{GT, c}|
\end{equation}
Higher $Acc_H$ means greater faithfulness. This aligns with intrinsic hallucination.

\textbf{4.4.2. Factor Utilization Recall:}
Measures incorporation of relevant provided factual information.
Factor Utilization Recall ($Rec_U$) is calculated as:
\begin{equation}
Rec_U = \left(\frac{N_{util}}{N_{gt}}\right) \times 100\%
\end{equation}
where $N_{util}$ is the count of utilized ground-truth factors, defined as:
\begin{equation}
N_{util} = \sum_{c \in \{c1, c2, c3\}} |F_{Ext, c} \cap F_{GT, c}|
\end{equation}
and $N_{gt}$ is the total count of ground-truth factors, as defined in the context of hallucination accuracy.
Higher $Rec_U$ means more comprehensive use of provided facts. This is akin to evaluating information usage and coverage. For successful abstentions, $Rec_U$ is 0.

\textbf{4.4.3. Successful Abstention Ratio:}
Evaluates ability to correctly refrain from generating arguments in ``mismatched'' or ``non-arguable'' scenarios.
The Successful Abstention Ratio ($Ratio_{Abstain}$) is defined by:
\begin{equation}
Ratio_{Abstain} = \left(\frac{N_{sa}}{N_{ta}}\right) \times 100\%
\end{equation}
where $N_{sa}$ is the number of successful abstentions (i.e., correct ``TERMINATE'' outputs), and $N_{ta}$ is the total number of case triples designed for abstention (i.e., instances from ``mismatched'' or ``non-arguable'' scenarios).
This is crucial for system reliability.

\section{Results and Analysis}
This section presents a detailed empirical analysis of the Reflective Multi-Agent framework's performance in comparison to the Single Agent, Single Agent with Enhanced Prompting, and Multi-Agent Debate without Reflection baselines. The evaluation is conducted across four distinct LLMs and three specifically designed legal scenarios: ``arguable,'' ``mismatched,'' and ``non-arguable.'' Performance is assessed using three key metrics: hallucination accuracy, factor utilization recall, and successful abstention ratio, each addressing aspects of argument quality, factual grounding, and ethical AI behavior.

\subsection{Hallucination Accuracy}
To assess the models' ability to avoid generating factually incorrect or unsupported statements (intrinsic hallucination), we measured hallucination accuracy. Table \ref{tab:hallucination_accuracy_results} presents the hallucination accuracy scores for each model and method across the three scenarios. Generally, all configurations achieve high accuracy (often >98\%) in ``arguable'' and ``mismatched'' scenarios. The ``non-arguable'' scenario, designed to lack a proper basis for argument, posed a more significant challenge to factual fidelity. The Reflective Multi-Agent setup consistently demonstrated superior or near-best performance, notably by substantially improving scores in the ``non-arguable'' scenario for all tested models. For instance, Reflective Multi-Agent with GPT-4o achieved 94.63\% accuracy in this challenging scenario, a marked improvement. This pattern suggests that when no legitimate argument can be formed, methods lacking robust reflective mechanisms are more prone to fabricating factors to fulfill the generation task—a tendency that the Reflective Multi-Agent framework effectively mitigates.

\begin{table*}[!htbp]
\caption{Hallucination Accuracy (\%) across Models, Methods, and Scenarios}
\label{tab:hallucination_accuracy_results}
\centering
\footnotesize
\setlength{\tabcolsep}{3.5pt} 
\begin{tabular}{llccc}
\toprule
\textbf{Model} & \textbf{Method} & \textbf{Arguable} & \textbf{Mismatched} & \textbf{Non-Arguable} \\
\midrule
\multirow{4}{*}{GPT-4o}
& Single Agent & 99.26 & 99.15 & 87.63 \\
& Single Agent with Enhanced Prompting & 99.32 & 99.20 & 85.96 \\
& Multi-Agent Debate without Reflection & 98.70 & 98.92 & 87.24 \\
& Reflective Multi-Agent & \textbf{99.81} & \textbf{99.44} & \textbf{94.63} \\
\midrule
\multirow{4}{*}{GPT-4o-mini}
& Single Agent & \textbf{99.33} & 95.75 & 78.06 \\
& Single Agent with Enhanced Prompting & 99.30 & 98.61 & 88.88 \\
& Multi-Agent Debate without Reflection & 99.03 & 98.12 & 83.17 \\
& Reflective Multi-Agent & 98.66 & \textbf{100.00} & \textbf{91.72} \\
\midrule
\multirow{4}{*}{Llama-4-Maverick-17b-128e}
& Single Agent & 99.59 & 96.88 & 93.69 \\
& Single Agent with Enhanced Prompting & 99.14 & 95.41 & 91.30 \\
& Multi-Agent Debate without Reflection & 99.33 & \textbf{99.14} & \textbf{97.67} \\
& Reflective Multi-Agent & \textbf{99.87} & 98.92 & 96.47 \\
\midrule
\multirow{4}{*}{Llama-4-Scout-17b-16e}
& Single Agent & 98.38 & 97.32 & 90.99 \\
& Single Agent with Enhanced Prompting & 98.75 & 98.19 & 87.79 \\
& Multi-Agent Debate without Reflection & 99.06 & 98.92 & 91.83 \\
& Reflective Multi-Agent & \textbf{99.26} & \textbf{99.91} & \textbf{98.14} \\
\bottomrule
\end{tabular}
\begin{minipage}{\textwidth}
\scriptsize Note: For each model, the highest value among methods for each scenario is bolded.
\end{minipage}
\end{table*}

The results in Table \ref{tab:hallucination_accuracy_results} confirms that hallucination is more prevalent in the ``non-arguable'' scenario. The Reflective Multi-Agent method generally shows high hallucination accuracy, with improvements in the ``non-arguable'' scenario. This indicates that the reflective process helps prevent models from fabricating facts when no valid basis exists.

\subsection{Factor Utilization Recall}
Effective legal arguments comprehensively incorporate relevant factual information from the provided case materials. Factor Utilization Recall was employed to measure the extent to which each method successfully utilized the ground-truth factors in the ``Arguable'' scenario, where substantive argument generation is expected. The results, detailed in Table \ref{tab:factor_utilization_recall_results}, indicate that multi-agent configurations (Multi-Agent and Reflective Multi-Agent) generally outperformed single-agent approaches (Single Agent and Single Agent with Enhanced Prompting). The introduction of reflection within the Reflective Multi-Agent framework provided a further discernible improvement over the Multi-Agent baseline. This suggests that both the interactive nature of multi-agent debate and the explicit analytical step of factor usage review contributed to a more thorough incorporation of case factors when arguments were contextually appropriate.

\begin{table*}[!htbp]
\caption{Factor Utilization Recall (\%) across Models and Methods (Arguable Scenario)}
\label{tab:factor_utilization_recall_results}
\centering
\footnotesize
\setlength{\tabcolsep}{3.5pt}
\begin{tabular}{llc}
\toprule
\textbf{Model} & \textbf{Method} & \textbf{Arguable} \\
\midrule
\multirow{4}{*}{GPT-4o}
& Single Agent & 87.47 \\
& Single Agent with Enhanced Prompting & 86.87 \\
& Multi-Agent Debate without Reflection & 89.55 \\
& Reflective Multi-Agent & \textbf{90.30} \\
\midrule
\multirow{4}{*}{GPT-4o-mini}
& Single Agent & 76.60 \\
& Single Agent with Enhanced Prompting & 67.39 \\
& Multi-Agent Debate without Reflection & 82.15 \\
& Reflective Multi-Agent & \textbf{88.58} \\
\midrule
\multirow{4}{*}{Llama-4-Maverick-17b-128e}
& Single Agent & 93.29 \\
& Single Agent with Enhanced Prompting & 87.81 \\
& Multi-Agent Debate without Reflection & 93.90 \\
& Reflective Multi-Agent & \textbf{94.18} \\
\midrule
\multirow{4}{*}{Llama-4-Scout-17b-16e}
& Single Agent & 90.02 \\
& Single Agent with Enhanced Prompting & 91.08 \\
& Multi-Agent Debate without Reflection & 95.76 \\
& Reflective Multi-Agent & \textbf{96.51} \\
\bottomrule
\end{tabular}
\begin{minipage}{\textwidth}
\scriptsize Note: For each model, the highest value among methods for the ``Arguable'' scenario is bolded. Factor Utilization Recall for ``Mismatched'' and ``Non-Arguable'' scenarios is not shown here as successful abstention (detailed in Table \ref{tab:successful_abstention_ratio_results}) is the primary success metric for those scenarios, making recall less informative.
\end{minipage}
\end{table*}

Table \ref{tab:factor_utilization_recall_results} indicates that Reflective Multi-Agent achieved high factor utilization in ``arguable'' scenarios. The removal of ``mismatched'' and ``non-arguable'' columns is because successful abstention, as shown in Table \ref{tab:successful_abstention_ratio_results}, is the more relevant metric for those scenarios.

\subsection{Successful Abstention Ratio}
One characteristic of a reliable and ethically sound AI system is its ability to recognize when an argument is untenable and to appropriately refrain from generating a response. The Successful Abstention Ratio evaluates this capacity, particularly in the ``mismatched'' and ``non-arguable'' scenarios where abstention is the desired outcome. As shown in Table \ref{tab:successful_abstention_ratio_results}, the Reflective Multi-Agent configuration demonstrated vastly superior performance in achieving successful abstention compared to all baseline methods.

\begin{table*}[!htbp]
\caption{Successful Abstention Ratio (\%) across Models, Methods (Mismatched \& Non-Arguable Scenarios)}
\label{tab:successful_abstention_ratio_results}
\centering
\footnotesize
\setlength{\tabcolsep}{3.5pt}
\begin{tabular}{llcc}
\toprule
\textbf{Model} & \textbf{Method} & \textbf{Mismatched} & \textbf{Non-Arguable} \\
\midrule
\multirow{4}{*}{GPT-4o}
& Single Agent & 0.00 & 0.00 \\
& Single Agent with Enhanced Prompting & 8.89 & 0.00 \\
& Multi-Agent Debate without Reflection & 0.00 & 0.00 \\
& Reflective Multi-Agent & \textbf{73.33} & \textbf{13.33} \\
\midrule
\multirow{4}{*}{GPT-4o-mini}
& Single Agent & 0.00 & 0.00 \\
& Single Agent with Enhanced Prompting & 1.11 & 17.78 \\
& Multi-Agent Debate without Reflection & 0.00 & 0.00 \\
& Reflective Multi-Agent & \textbf{92.22} & \textbf{26.67} \\
\midrule
\multirow{4}{*}{Llama-4-Maverick-17b-128e}
& Single Agent & 0.00 & 5.56 \\
& Single Agent with Enhanced Prompting & 8.89 & 11.11 \\
& Multi-Agent Debate without Reflection & 1.11 & 6.67 \\
& Reflective Multi-Agent & \textbf{87.78} & \textbf{87.78} \\
\midrule
\multirow{4}{*}{Llama-4-Scout-17b-16e}
& Single Agent & 1.11 & 0.00 \\
& Single Agent with Enhanced Prompting & 2.22 & 0.00 \\
& Multi-Agent Debate without Reflection & 1.11 & 0.00 \\
& Reflective Multi-Agent & \textbf{92.22} & \textbf{42.22} \\
\bottomrule
\end{tabular}
\begin{minipage}{\textwidth}
\scriptsize Note: For each model, the highest value among methods for each scenario is bolded.
\end{minipage}
\end{table*}

Table \ref{tab:successful_abstention_ratio_results} underscores the strength of the Reflective Multi-Agent framework in this domain. While Single Agent with Enhanced Prompting and Multi-Agent configurations offered some marginal improvements over the basic Single Agent method, their abstention capabilities remained limited. In contrast, Reflective Multi-Agent consistently achieved higher successful abstention rates across all tested LLMs. This finding suggests that the Factor Analyst agent, with its explicit mandate to assess the groundability of an argument and trigger termination if necessary, played an effective role as a gatekeeper against the generation of inappropriate or unsupportable claims.

\subsection{Synthesis of Findings}
The results from the evaluation of hallucination accuracy, factor utilization recall, and successful abstention ratio revealed a synergistic relationship between the multi-agent architecture and the reflection mechanisms embedded within the Reflective Multi-Agent framework. The Multi-Agent setup demonstrated benefits over Single Agent approaches, particularly in enhancing factor utilization, due to the dynamic of argument and counter-argument construction. However, it was the introduction of explicit, role-based reflection in Reflective Multi-Agent that elevated performance across the aspects related to ethical and reliable argument generation.

Reflective Multi-Agent's superiority was most pronounced in reducing hallucinations, especially in ``non-arguable'' contexts where the temptation for models to invent information was high, and in achieving successful abstention when arguments were ungroundable. Importantly, this improvement in successful abstention did not come at the cost of performance in other key areas; the Reflective Multi-Agent framework generally maintained or enhanced hallucination accuracy (Table \ref{tab:hallucination_accuracy_results}) and factor utilization recall (Table \ref{tab:factor_utilization_recall_results}) compared to baselines. The Factor Analyst component acted as a dedicated mechanism for ensuring factual grounding and prompting abstention when warranted. The Argument Polisher, while not directly measured by these quantitative metrics, contributed by refining the output based on the Factor Analyst's feedback, ensuring that any necessary revisions maintained factual integrity. While multi-agent debate contributed to exploring what could be argued, the reflective layer in Reflective Multi-Agent controlled how and whether to argue. This combination allowed the Reflective Multi-Agent framework to not only generate more comprehensive and factually sound arguments when appropriate but also to reliably abstain when arguments were untenable, thereby addressing key dimensions of the research questions (RQ1 and RQ2) regarding performance and the impact of reflection. Consequently, these improvements in generating well-grounded arguments and appropriately abstaining contribute to the development of more ethically persuasive and less manipulative legally compliant intelligent chatbots, a key concern of RQ3.

\section{Discussion}
\subsection{Technical Performance and Ethical Goals}
The quantitative improvements by Reflective Multi-Agent in hallucination accuracy, factor utilization, and abstention directly corresponded to developing a more ethically sound and less manipulative AI for legal applications.

\textbf{Factor Utilization as Transparent Persuasion:} High factor utilization meant arguments were based on provided evidence, making reasoning transparent. Ethical persuasion is evidence-based, aligning with legal principles requiring substantiated arguments. Comprehensive factor utilization was key to persuasive arguments \cite{nadler1983evidence}.

\textbf{Hallucination Accuracy and Abstention as Safeguards Against Manipulation:} Generating statements on fabricated factors was informational manipulation \cite{rhode1985ethical}. Reflective Multi-Agent's superior hallucination accuracy, especially in ``non-arguable'' scenarios, curtailed this. Attempting arguments in ``mismatched'' or ``non-arguable'' scenarios misled users. Reflective Multi-Agent's high abstention rates safeguarded against this, ensuring the system knew when to stop, crucial for responsible AI.

\subsection{Error Analysis: Illustrative Examples of RMA Superiority}
Consider these illustrative examples:

\begin{custombox}{Example 1: Hallucination in an Arguable Scenario}
    \textbf{Input Case Triple:}\\
    c1 (Current Case): \\{F1 Disclosure-in-negotiations (D), F3 Employee-sole-developer (D), F6 Security-measures (P), F20 Info-known-to-competitors (D), F25 Info-reverse-engineered (D)\\}
    c2 (Plaintiff's Case): \\{outcome Plaintiff\\} {F3 Employee-sole-developer (D), F6 Security-measures (P), F11 Vertical-knowledge (D), F12 Outsider-disclosures-restricted (P), F14 Restricted-materials-used (P), F20 Info-known-to-competitors (D)\\}
    c3 (Defendant's Case): \\{outcome Defendant\\} {F3 Employee-sole-developer (D), F6 Security-measures (P), F10 Secrets-disclosed-outsiders (D), F16 Info-reverse-engineerable (D), F25 Info-reverse-engineered (D)\\} \\
    \textbf{Model:} \texttt{Llama-4-Maverick-17b-128e}.\\ \\
    \textbf{Single Agent with Enhanced Prompting Output (Plaintiff's Rebuttal):} ``... Moreover, the input case has F12 Outsider-disclosures-restricted (P) and F14 Restricted-materials-used (P) supporting the trade secret claim through the implication of robust protective measures ...''\\ \\
    \textit{Error:} Single Agent with Enhanced Prompting misattributed F12 and F14 to c1.\\ \\
    \textbf{Reflective Multi-Agent Output (Plaintiff's Rebuttal):} ``... The presence of F6 Security-measures (P) in both TSC1 and the input case underscores the Plaintiff's efforts to safeguard their trade secrets, a crucial element in trade secret misappropriation claims ...''\\ \\
    \textit{RMA Process:} Factor Analyst identified factor misattribution and requested revision.\\ \\ 
    \textit{Superiority:} Reflective Multi-Agent correctly developed a grounded analogy.
\end{custombox}

\begin{custombox}{Example 2: Failure to Abstain in a Non-Arguable Scenario}
    \textbf{Input Case Triple:}\\
    c1 (Current Case): \\{F6 Security-measures (P), F15 Unique-product (P), F18 Identical-products (P), F20 Info-known-to-competitors (D)\\}
    c2 (Plaintiff's Case): \\{outcome Plaintiff\\} {F7 Brought-tools (P), F10 Secrets-disclosed-outsiders (D), F17 Info-independently-generated (D), F23 Waiver-of-confidentiality (D), F27 Disclosure-in-public-forum (D)\\}
    c3 (Defendant's Case): \\{outcome Defendant\\} {F3 Employee-sole-developer (D), F8 Competitive-advantage (P), F10 Secrets-disclosed-outsiders (D), F22 Invasive-techniques (P), F24 Info-obtainable-elsewhere (D), F27 Disclosure-in-public-forum (D)\\} \\
    \textbf{Model:} \texttt{GPT-4o-mini}.\\ \\
    \textbf{Single Agent with Enhanced Prompting Output (Defendant's Argument):} ``... c3 (outcome Defendant) is analogous; both (c1 and c3) share F10 Secrets-disclosed-outsiders (D) and F27 Disclosure-in-public-forum (D), indicating a lack of confidentiality and potential public knowledge of the information ...''\\ \\
    \textit{Error:} Single Agent with Enhanced Prompting forced an argument and claimed ``F10 and F27'' in c1, creating a manipulative and ungrounded analogy.\\ \\
    \textbf{Reflective Multi-Agent Output (Defendant's Argument):} ``TERMINATE: Generation stopped. The argument must be abstained from because there are no common factors between the input case and the primary cited c3. None of the current case's factors are present in c3.''\\ \\
    \textit{RMA Process:} Factor Analyst identified a lack of grounding and mandated ``TERMINATE''.\\ \\ 
    \textit{Superiority:} Reflective Multi-Agent correctly abstained.
\end{custombox}

These examples show how Reflective Multi-Agent's structured reflection contributed to more accurate, reliable, and ethically sound outputs.

\section{Limitations and Future Work}

The Reflective Multi-Agent framework promoted ethical persuasion and mitigated manipulation. Reflective Multi-Agent's evidence-grounded arguments fostered transparency. Enhanced hallucination accuracy and abstention reduced misleading users \cite{tonmoy2024comprehensive}. Legal professionals have duties of competence and candor with AI \cite{davis2020future}. Reflective Multi-Agent could assist in meeting these. However, AI perpetuating biases in legal data remained a challenge \cite{draper2023potential}.

\subsection{Limitations of the Current Research}
The current research, while demonstrating the promise of the Reflective Multi-Agent framework, was subject to several limitations that warranted discussion. A primary constraint was the scope of legal factors; the study employed a predefined, discrete set of factors. Real-world legal reasoning, however, is considerably more nuanced and often involved interpreting and weighing factors that were not explicitly enumerated. Consequently, the factor definition and extraction stages, treated here as a pre-process, represented a simplification of a complex interpretative task. Another limitation was the depth of reflection embedded in the Reflective Multi-Agent framework. The current implementation involved a single round of analysis and polishing by the Factor Analyst and Argument Polisher, respectively, with at most one revision cycle. While effective, more sophisticated or iterative reflection mechanisms might yield further improvements in argument quality, though this would likely come at an increased computational cost and require careful design to avoid excessive processing times.

Furthermore, the performance of the Reflective Multi-Agent framework was inherently bounded by the capabilities of the underlying LLMs used for the agent roles. Any inherent biases, knowledge gaps, or reasoning limitations of these base models would invariably influence the final output, regardless of the structured reflection process. The evaluation metrics, while quantitative and objective, also served as proxies for the true, multifaceted quality of legal argumentation. Specifically, the use of an LLM-as-a-Judge for assessing factor utilization, while scalable, had potential for inherent biases or misinterpretations. Therefore, comprehensive human evaluation remained indispensable for a complete assessment. Finally, the current study's focus on a specific 3-ply argument generation task within U.S. trade secret law meant that the scalability and generalizability of the Reflective Multi-Agent framework to wider legal domains, different jurisprudential contexts, and more complex argument structures still needed thorough assessment.

\subsection{Directions for Future Research}
Building upon the findings and limitations of this work, several promising directions for future research emerged. A key area for advancement was through enhanced factor analysis and knowledge grounding. This could involve developing methods for the system to identify implicit factors not explicitly listed in the input, assess the dynamic relevance of factors based on evolving case narratives, or integrate Retrieval Augmented Generation (RAG) techniques to ground arguments in external legal knowledge bases such as statutes, case law repositories, or scholarly articles \cite{zheng2025reasoning}. Such enhancements would move beyond reliance on pre-defined factor sets and allow for more robust and contextually aware reasoning. While the current study demonstrated strong performance using factor-represented inputs, future work should explore the adaptation of the RMA framework to process full-text legal documents directly. This would involve tackling challenges in automated information extraction but could unlock the potential to capture richer contextual nuances from unstructured legal narratives. Another important avenue involved advanced argument polishing and rhetorical control. This could entail fine-tuning the Argument Polisher agent on specialized corpora of legal writing to better capture an appropriate legal style and tone, or developing mechanisms that allowed users to explicitly yet ethically guide the rhetorical strategies employed in the generated arguments \cite{wu2023autogen,sun2024llm}, ensuring that persuasiveness did not compromise fairness or accuracy.

Further research should also explore the development of dynamic and adaptive reflection mechanisms. Instead of a fixed number of reflection cycles, the system could be designed to adjust the depth or intensity of reflection based on the initial quality of the generated argument, the complexity of the legal scenario, or even the confidence scores of the agents involved. Designing effective human-in-the-loop collaboration systems was another direction, allowing legal professionals to interact with, guide, and iteratively refine the arguments generated by the Reflective Multi-Agent framework. Further effort must also be dedicated to addressing potential biases, both those that might be present in the input case factors and those that could be introduced or amplified by the LLMs during the generation process. Methodologies for bias detection and mitigation would be crucial for ensuring equitable and just outcomes. Exploring diverse multi-agent configurations, such as introducing an adversarial agent to proactively challenge assertions or experimenting with different numbers of agents and interaction protocols \cite{han2024llm}, could also lead to more robust arguments. Finally, longitudinal studies on the trust and reliance placed on such AI systems by legal professionals were essential to understand their real-world impact, identify potential risks of over-reliance or misuse, and develop guidelines for their responsible deployment in legal practice.

\section{Conclusion}
This paper introduced and evaluated a Reflective Multi-Agent framework to improve LLM-generated 3-ply legal arguments, focusing on ethical persuasion and mitigating manipulation. Empirical results showed Reflective Multi-Agent framework's advantages over baselines.
Reflective Multi-Agent framework achieved vastly superior successful abstention, crucial for preventing misleading discourse. It markedly improved hallucination accuracy (especially in non-arguable contexts) and enhanced factor utilization recall.
These findings suggest structured reflection (Factor Analyst, Argument Polisher) within a multi-agent debate offered a robust method for guiding LLMs towards more reliable, ethically sound outputs. By analyzing arguments for grounding, factor use, abstention necessity, and polishing for clarity, Reflective Multi-Agent addressed key LLM weaknesses in law.
Developing such systems was important for trustworthy legally compliant intelligent chatbots. While challenges remained, Reflective Multi-Agent's principles (role specialization, iterative refinement, explicit analysis) provided a promising direction for AI in law, fostering systems that were persuasive, responsible, and ethical.

\section*{Declaration of Use of AI}
The authors declare that no artificial intelligence tools were used in the writing of this manuscript.

\printbibliography

\appendix

\section{Detailed Prompts}
\label{appendix:prompts}

This appendix provided examples of the prompts used for different components of the system.

\begin{custombox}{Prompt for Factor Analyst}
    \label{prompt:factor_analyst}
You are a Factor Analyst Agent. Your task is to analyze a given legal argument segment against c1 and relevant Trade Secret Case(s) (c2, c3). Your goal is to determine if the argument segment must be abstained from, if it requires correction of factual errors, or if it appears valid based on the provided data.

\textbf{IMPORTANT CONTEXT:} You will be analyzing one ply of a 3-ply argument at a time (Plaintiff's Argument using c2, Defendant's Counterargument using c3, or Plaintiff's Rebuttal addressing c3 and reinforcing with c2). Pay close attention to which party is arguing and which precedent (c2 or c3) is primarily being cited or addressed in the segment provided.

\textbf{Follow this process STRICTLY:}
\begin{enumerate}
    \item \textbf{Identify the Context:}
    \begin{itemize}
        \item Determine the party making the argument segment (Plaintiff or Defendant).
        \item Identify the primary precedent being cited or addressed in this segment (c2 for Plaintiff's Argument, c3 for Defendant's Counterargument, c2 \& c3 for Plaintiff's Rebuttal). You will be given the factors and outcome for the relevant precedent(s).
    \end{itemize}
    \item \textbf{Determine if Abstention is REQUIRED (This is the FIRST and most critical check):}
    \begin{itemize}
        \item \textbf{Focus ONLY on the PRIMARY cited precedent} for the argument ply being evaluated (c2 for Plaintiff's Arg, c3 for Defendant's Counter). For Rebuttal, consider the check against c2 for reinforcement.
        \item Verify the actual common factors between c1 and this \textit{primary} cited precedent (c2 or c3) based \textit{only} on the provided factor lists. Ignore factors mentioned from other, non-primary precedents during this step.
        \item Check the actual outcome of the \textit{primary} cited precedent and whether it favors the arguing party (Plaintiff needs Plaintiff-outcome c2, Defendant needs Defendant-outcome c3).
        \item \textbf{Abstention IS REQUIRED} and you MUST output \texttt{``REQUIRES\_ABSTENTION''} if EITHER of the following conditions is true for the primary cited precedent:
        \begin{enumerate}
            \item There are ZERO genuinely common factors between c1 and the \textit{primary} cited precedent (c2 or c3) used for the core analogy/argument. (For Rebuttal, check this specifically for factors cited from c2 for reinforcement). Count common factors carefully. If the count is 0, abstention is mandatory.
            OR
            \item The actual outcome of the \textit{primary} cited precedent is UNFAVORABLE to the party making this argument segment (e.g., Plaintiff citing a Defendant-outcome c2, Defendant citing a Plaintiff-outcome c3).
        \end{enumerate}
        \item If abstention is required, set analysis\_outcome to \texttt{``REQUIRES\_ABSTENTION''}, provide the reason, and proceed to output formatting. DO NOT proceed to step 3.
    \end{itemize}
    \item \textbf{If Abstention is NOT Required, then Determine if Correction is Needed:}
    \begin{itemize}
        \item Identify all factors claimed as common or distinguishing in the argument segment. Compare these against the actual factor lists for c1 and \textit{all} relevant precedents (c2 \& c3).
        \item Identify if the argument segment misrepresents the outcome of \textit{any} cited precedent.
        \item \textbf{Correction IS REQUIRED} if:
        \begin{enumerate}
            \item The argument claims common factors that are fabricated (e.g., a factor claimed as common between c1 and precedentX is not present in both's actual lists).
            \item The argument claims distinguishing factors that are fabricated (e.g., claiming precedentX has factor Y which it doesn't, or claiming c1 lacks factor Z which it has).
            \item The argument misrepresents the actual outcome of a cited precedent (and this wasn't caught by the abstention rule).
        \end{enumerate}
        \item If correction is required (and abstention was not), your analysis\_outcome is \texttt{``REQUIRES CORRECTION''}. List the specific errors.
    \end{itemize}
    \item \textbf{If Neither Abstention nor Correction is Required:}
    \begin{itemize}
        \item Your analysis\_outcome is \texttt{``VALID ARGUMENT''}.
    \end{itemize}
\end{enumerate}

\textbf{Special Notes for Plaintiff's Rebuttal:}
\begin{itemize}
    \item The Rebuttal aims to distinguish c3 (cited by Defendant) and reinforce the Plaintiff's case (potentially citing c2 again).
    \item Analyze claims about c3: Are the claimed distinguishing factors accurate based on the actual factor lists of c1 and c3?
    \item Analyze claims about c2 (if used for reinforcement): Are the claimed common factors accurate? Is the outcome still favorable?
    \item Abstention (Rule 2a) applies if the reinforcement part \textit{claims} common factors with c2 but there are actually zero \textit{and} no valid distinction of c3 is made. Abstention (Rule 2b) applies if c2 (used for reinforcement) has an unfavorable outcome.
    \item Correction (Rule 3) applies if \textit{any} factor claims (common or distinguishing, regarding c2 or c3) are fabricated or misrepresented.
\end{itemize}

Output your analysis in JSON format as specified below. Ensure the JSON is the only output.

\textbf{JSON Output Format:}
\texttt{\{}
  \texttt{"analysis\_outcome": "REQUIRES\_ABSTENTION" / "REQUIRES\_CORRECTION" / "VALID\_ARGUMENT",}
  \texttt{"summary": "A concise explanation. If abstention, state the specific reason (unfavorable outcome OR zero common factors for the *primary* cited precedent). If correction, summarize key factual errors. If valid, confirm.",}
  \texttt{"abstention\_details": \{ // Include this section ONLY if analysis\_outcome is "REQUIRES\_ABSTENTION"}
    \texttt{"reason\_for\_abstention": "No common factors found." / "Cited precedent outcome is unfavorable for the arguing party." / "Both: No common factors and unfavorable precedent outcome."}
  \texttt{\},}
  \texttt{"correction\_details": \{ // Include this section ONLY if analysis\_outcome is "REQUIRES\_CORRECTION"}
    \texttt{"fabricated\_or\_misrepresented\_factors": ["Factor A (P) - claimed as common but not in c2's actual factors", "Factor B (D) - claimed for c1 but not present in c1's actual factors"], // List factors that are incorrectly claimed. Be specific about the error.}
    \texttt{"misrepresented\_tsc\_outcome": "e.g., Argument claims c2 outcome is Plaintiff, but actual outcome is Defendant.", // Describe if precedent outcome is misrepresented. Omit or null if not applicable. (Note: "tsc" kept in key for consistency with original prompt key, value changed to precedent)}
    \texttt{"other\_issues\_for\_correction": "Brief description of any other critical factual errors needing correction." // Omit or null if not applicable.}
  \texttt{\}}
\texttt{\}}

\textbf{Example 1 (Requires Abstention - unfavorable outcome):}
Argument: Plaintiff's argument cites c2. Provided data: c2 actual outcome is 'Defendant'.
Output:
\texttt{\{}
  \texttt{"analysis\_outcome": "REQUIRES\_ABSTENTION",}
  \texttt{"summary": "The argument for Plaintiff, citing c2, must be abstained from. c2's actual outcome is 'Defendant', which does not favor the Plaintiff.",}
  \texttt{"abstention\_details": \{}
    \texttt{"reason\_for\_abstention": "Cited precedent outcome is unfavorable for the arguing party."}
  \texttt{\}}
\texttt{\}}

\textbf{Example 2 (Requires Abstention - no common factors):}
Argument: Cites precedentX. Provided data: c1 factors \{F1, F2\}, precedentX factors \{F3, F4\}. (No common factors).
Output:
\texttt{\{}
  \texttt{"analysis\_outcome": "REQUIRES\_ABSTENTION",}
  \texttt{"summary": "The argument must be abstained from as there are no common factors between c1 and the cited precedentX.",}
  \texttt{"abstention\_details": \{}
    \texttt{"reason\_for\_abstention": "No common factors found."}
  \texttt{\}}
\texttt{\}}

\textbf{Example 3 (Requires Correction - fabricated factor):}
Argument for Plaintiff cites c2. Provided data: c2 actual outcome 'Plaintiff'. c1 \{F1, F2\}, c2 \{F1, F3\}.
Argument claims: "c1 and c2 share F1 and F4." (F4 is fabricated as it's not in c2 and not common).
Output:
\texttt{\{}
  \texttt{"analysis\_outcome": "REQUIRES\_CORRECTION",}
  \texttt{"summary": "The argument requires correction. Factor F4 was claimed as common with c2, but F4 is not present in c2's actual factors.",}
  \texttt{"correction\_details": \{}
    \texttt{"fabricated\_or\_misrepresented\_factors": ["F4 (claimed as common with c2 but not present in c2's actual factors)"]}
  \texttt{\}}
\texttt{\}}

\textbf{Example 4 (Valid Argument):}
Argument for Plaintiff cites c2. Provided data: c2 actual outcome 'Plaintiff'. c1 \{F1, F2\}, c2 \{F1, F3\}.
Argument claims: "c1 and c2 share F1."
Output:
\texttt{\{}
  \texttt{"analysis\_outcome": "VALID\_ARGUMENT",}
  \texttt{"summary": "The argument segment appears valid. The cited precedent outcome favors the arguing party, and the claimed common factor (F1) is verified."}
\texttt{\}}
\end{custombox}

\begin{custombox}{Prompt for Argument Polisher}
    \label{prompt:argument_polisher}
You are an Argument Polisher Agent. You will receive a generated legal argument segment, the original c1 factors, relevant c2/c3 factors, and the Factor Analyst's report.
Your tasks:
\begin{enumerate}
    \item Review the argument segment for factual accuracy based on the provided case factors and Factor Analyst's report.
    \item Check for logical coherence and persuasive strength. Specifically, assess factor utilization:
    \begin{enumerate}
        \item Are all relevant supporting factors from c1 and cited precedent (c2/c3) effectively used to build the analogy or argument?
        \item Are distinguishing factors (both in the cited precedent (c2/c3) not present in c1, and in c1 not present in the cited precedent (c2/c3)) clearly highlighted when making distinctions or counterarguments?
        \item Are there any crucial factors from c1 or precedents (c2/c3) that have been overlooked and could strengthen or weaken the argument?
    \end{enumerate}
    \item Provide feedback on inaccuracies, argument strength, and specifically on factor utilization.
    \item If revisions are needed, provide clear instructions to the Argument Developer Agent on what to correct or improve, with a strong focus on enhancing factor utilization.
\end{enumerate}

Output your assessment in JSON format:
\texttt{\{}
  \texttt{"argument\_segment\_type": "Plaintiff's Argument / Defendant's Counterargument / Plaintiff's Rebuttal",}
  \texttt{"accuracy\_assessment": "Accurate / Minor Inaccuracies / Major Inaccuracies",}
  \texttt{"strength\_assessment": "Strong / Moderate / Weak (based on factor utilization and logic)",}
  \texttt{"factor\_utilization\_assessment": "Excellent / Good / Fair / Poor",}
  \texttt{"feedback\_summary": "e.g., 'The argument correctly identifies shared factors but misses a key distinguishing factor in c2. Factor utilization could be improved by incorporating F\_X from c1.'",}
  \texttt{"revision\_needed": true/false,}
  \texttt{"instructions\_for\_developer": "If revision\_needed is true, provide concise instructions. e.g., 'Re-evaluate c2. While F4 is common, c2 also has F7 (P) which is a key distinction you missed. c1 has F10 (D) which weakens your analogy. Strengthen your argument by explicitly mentioning how F10 (D) is overcome or why c2 is still a good precedent despite it. Ensure all favorable factors for your side common to c1 and c2 are mentioned.'"}
\texttt{\}}
\end{custombox}

\begin{custombox}{Core Prompt Structure for 3-Ply Argument Generation}
    \label{prompt:ts_misappropriation_appendix} 
You are an AI assistant tasked with formulating legal arguments for trade secret misappropriation claims.
Construct a 3-Ply Argument:
\begin{enumerate}
    \item Plaintiff's Argument: Cite a relevant Trade Secret Case (c2) with a favorable outcome for Plaintiff. Highlight shared factors between c1 and c2.
    \item Defendant's Counterargument: Distinguish c2. Cite a counterexample (c3, with a Defendant-favorable outcome) and draw an analogy to c1, highlighting shared factors between c1 and c3.
    \item Plaintiff's Rebuttal: Address and distinguish c3, reinforcing the Plaintiff's original argument (e.g. by re-emphasizing shared factors between c1 and c2, or distinguishing c1 from c3 on further grounds).
\end{enumerate}

Base your arguments on the provided factors. Ensure logical consistency.
Output the 3-ply argument in a single JSON object with keys: \texttt{``Plaintiff's Argument''}, \texttt{``Defendant's Counterargument''}, \texttt{``Plaintiff's Rebuttal''}.

\textbf{Example c1:}
\texttt{F1 Disclosure-in-negotiations (D)}\\
\texttt{F4 Agreed-not-to-disclose (P)}\\
\texttt{F6 Security-measures (P)}

\textbf{Example c2 (for Plaintiff):}
outcome Plaintiff\\
\texttt{F4 Agreed-not-to-disclose (P)}\\
\texttt{F6 Security-measures (P)}\\
\texttt{F7 Brought-tools (P)}

\textbf{Example c3 (for Defendant):}
outcome Defendant\\
\texttt{F1 Disclosure-in-negotiations (D)}\\
\texttt{F5 Agreement-not-specific (D)}

\textbf{Example JSON Output:}
\texttt{\{}
  \texttt{"Plaintiff's Argument": "Factors F4 Agreed-not-to-disclose (P) and F6 Security-measures (P) were present in both c1 and c2 (outcome Plaintiff), supporting the Plaintiff. c1 also has F12...",}
  \texttt{"Defendant's Counterargument": "c2 is distinguishable because it had F7 Brought-tools (P), not in c1. Furthermore, c1 has F1 Disclosure-in-negotiations (D). c3 (outcome Defendant) is analogous; c1 and c3 share F1 Disclosure-in-negotiations (D) and F5 Agreement-not-specific (D).",}
  \texttt{"Plaintiff's Rebuttal": "c3 is distinguishable as c1 lacks F5 Agreement-not-specific (D) and has strong pro-plaintiff factors like F4 and F6 not in c3."}
\texttt{\}}

If you cannot make a valid argument for a step (e.g., no suitable precedent), state that clearly for that part of the argument.
\end{custombox}

\begin{custombox}{Prompt for Factor Extraction by External LLM}
    \label{prompt:factor_extraction_appendix}
You are a Factor Distiller Agent. Given a 3-ply legal argument in JSON format, extract all unique legal factors mentioned for \texttt{``c1''}, \texttt{``c2''}, and \texttt{``c3''}.
Factors are in the format like \texttt{``F1 Disclosure-in-negotiations (D)''}, \texttt{``F4 Agreed-not-to-disclose (P)''}, etc.
Output the results as a JSON object with keys \texttt{``c1''}, \texttt{``c2''}, and \texttt{``c3''}, where each value is a list of unique factor strings.

\textbf{Example Input Argument JSON:}
\texttt{\{}
  \texttt{"Plaintiff's Argument": "c1 shares F4 (P) and F6 (P) with c2 (outcome Plaintiff). c1 also features F12 (P).",}
  \texttt{"Defendant's Counterargument": "c2 also had F7 (P), distinguishing it. c1 has F1 (D). c3 (outcome Defendant) is similar, c1 and c3 share F1 (D).",}
  \texttt{"Plaintiff's Rebuttal": "c3 is different, c1 does not have F5 (D) which was in c3."}
\texttt{\}}

\textbf{Example Output JSON:}
\texttt{\{}
  \texttt{"c1": ["F4 Agreed-not-to-disclose (P)", "F6 Security-measures (P)", "F12 Outsider-disclosures-restricted (P)", "F1 Disclosure-in-negotiations (D)"],}
  \texttt{"c2": ["F4 Agreed-not-to-disclose (P)", "F6 Security-measures (P)", "F7 Brought-tools (P)"],}
  \texttt{"c3": ["F1 Disclosure-in-negotiations (D)", "F5 Agreement-not-specific (D)"]}
\texttt{\}}
Ensure each factor appears only once per list, even if mentioned multiple times in the argument.
\end{custombox}

\section{Case Triple Dataset Characteristics}
\label{appendix:dataset}
The dataset comprised synthetically generated factor-based case triples. Each triple included a current case (c1), a plaintiff precedent (c2), and a defendant precedent (c3). Three scenarios were created:
\begin{itemize}
    \item \textbf{Arguable:} c1 shares relevant factors with c2 (supporting Plaintiff) and c3 (supporting Defendant).
    \item \textbf{Mismatched:} Precedent outcomes conflict with the side they are meant to support.
    \item \textbf{Non-Arguable:} No relevant factor overlaps between c1 and c2, or c1 and c3.
\end{itemize}
These scenarios were designed to test substantive argument generation, recognition of contextual inappropriateness, and abstention capabilities, respectively. Refer to Table \ref{tab:case_set_examples} for illustrative examples.

\end{document}